\documentclass[sigconf]{acmart}

\usepackage{xcolor}
\usepackage{colortbl}
\usepackage{booktabs}
\usepackage{multirow}
\usepackage[normalem]{ulem}  

\definecolor{yilin}{RGB}{0, 90, 200}
\definecolor{lightgray}{gray}{0.90}

\newcommand{\first}[1]{\textbf{#1}}
\newcommand{\second}[1]{\underline{#1}}


\graphicspath{{figures/}{./}}

\AtBeginDocument{%
  }

\copyrightyear{2026}
\acmYear{2026}
\setcopyright{cc}
\setcctype{by}

\acmConference[MM '26]
  {Proceedings of the 34th ACM International Conference on Multimedia}
  {November 10--14, 2026}
  {Rio de Janeiro, Brazil}

\acmBooktitle{Proceedings of the 34th ACM International Conference on Multimedia
  (MM '26), November 10--14, 2026, Rio de Janeiro, Brazil}

\acmDOI{10.1145/3767308.3836583}
\acmISBN{979-8-4007-2213-4/2026/11}

\usepackage{enumerate}
\usepackage{enumitem}

\begin{document}

\title{Same Semantics, Different Paths: Self-Improving Alignment for Vision-Text Compression}

\author{Tianyu Liang}
\affiliation{%
  \institution{Southeast University}
  \city{Nanjing}
  \country{China}
}
\email{220245012@seu.edu.cn}

\author{Xiangxi Zheng}
\affiliation{%
  \institution{Nanjing University}
  \city{Nanjing}
  \country{China}
}
\email{zhengxx@smail.nju.edu.cn}

\author{Yilin Wang}
\affiliation{%
  \institution{Zhejiang University}
  \city{Hangzhou}
  \country{China}
}
\email{yilin.wang@zju.edu.cn}

\author{Dongxing Mao}
\correspondingauthor
\affiliation{%
  \institution{National University of Singapore}
  \city{Singapore}
  \country{Singapore}
}
\email{dongxing.mao@u.nus.edu}

\renewcommand{\shortauthors}{Liang et al.}

\begin{abstract}
Vision-Text Compression (VTC) renders long texts into images and encodes them through the vision encoder (ViT), compressing thousands of text tokens into far fewer visual tokens. However, since the ViT is pretrained predominantly on natural images, it captures visual attributes (glyphs, font sizes, layout) rather than linguistic semantics, causing rendered-image representations to diverge from native-text representations. We term this \emph{cross-path inconsistency} and show, via rendering perturbation experiments, that it is a critical yet overlooked bottleneck of VTC.
We propose \textbf{SPIRAL} (\textbf{S}elf-improving \textbf{P}ath \textbf{I}ntegration and \textbf{R}e\-alignment), a self-supervised alignment framework that closes this gap using only the model's own text-path behavior as supervision, requiring no external teachers or additional annotations. SPIRAL operates at two complementary granularities: token-level \emph{on-policy distillation} (OPD) for local faithfulness, and sequence-level \emph{preference optimization} (DPO) for global coherence.
On VTCBench, SPIRAL improves the overall score of Qwen3-VL-8B from \textbf{35.10} to \textbf{54.02}, approaching the native text-input performance (\textbf{55.60}) and outperforming models up to 30$\times$ larger. The two granularities exhibit complementary strengths: OPD excels at retrieval and is sample-efficient, while DPO is stronger on reasoning and memory and scales better with data. SPIRAL's benefits also generalize to out-of-domain benchmarks, confirming that effective VTC hinges on aligning rendered-image representations back to native-text semantics.
\end{abstract}




\maketitle

\section{Introduction}

\begin{figure}[t]
  \centering
  \includegraphics[width=\columnwidth]{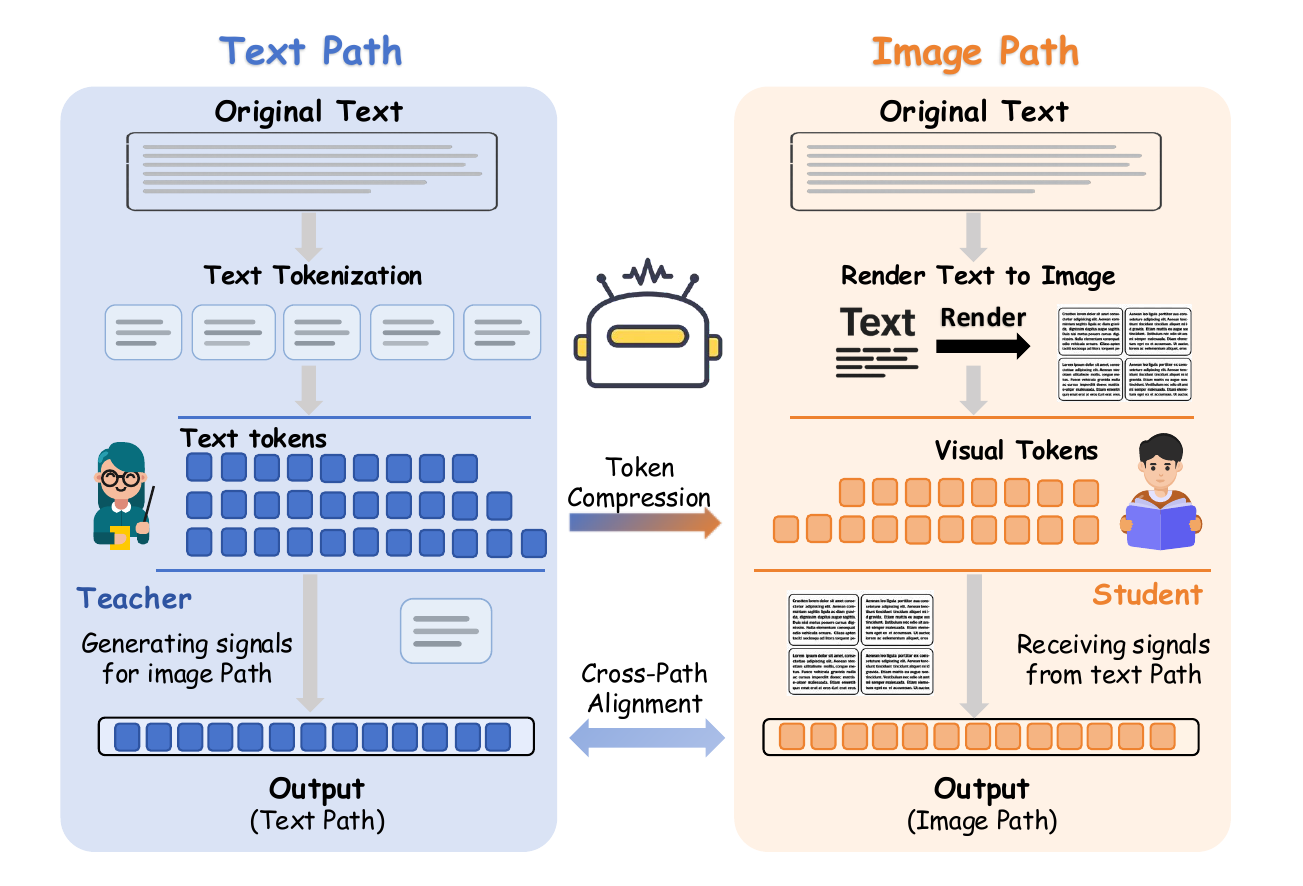}
  \caption{Overview of cross-path inconsistency in VTC. The same original text is processed through two paths: the \emph{text path} (left) tokenizes text natively, while the \emph{image path} (right) renders text into images and encodes them as visual tokens. Although both paths share the same MLLM backbone, the ViT encoder captures visual attributes rather than linguistic semantics, producing divergent representations. SPIRAL aligns the image path toward the text path using the model's own text-derived signals.}
  \Description{Diagram showing the text path and image path in VTC, with the text path generating supervision signals for the image path.}
  \label{fig:motivation}
\end{figure}

Large multimodal models (MLLMs) have achieved strong performance across visual understanding, document analysis, and cross-modal reasoning. Yet long-context modeling remains a critical bottleneck: attention complexity and KV cache memory grow sharply with input length~\cite{beltagy2020longformerlongdocumenttransformer,zaheer2020big,dao2022flashattention}, and existing text-side remedies such as sparse attention and KV cache compression provide only partial relief. Vision-Text Compression (VTC)~\cite{wei2025deepseekocrcontextsopticalcompression,cheng2025glyphscalingcontextwindows,li2025textorpixels} offers a fundamentally different solution: rendering long text into high-density images and encoding them via the MLLM's existing vision encoder (ViT)~\cite{wang2024visincontext,lu2024textpixeladvancinglongcontext,xing2025visioncentrictokencompressionlarge}, achieving up to $2.5\times$ token compression and $2\times$ inference speedup. However, we find that this promising approach harbors a critical yet overlooked problem.

Since the ViT is pretrained predominantly on natural images, it captures visual attributes (glyphs and layout) rather than linguistic semantics.  Consequently, even when the underlying text content is identical, representations from the \emph{image path} (text $\rightarrow$ render $\rightarrow$ ViT) diverge significantly from those of the \emph{text path} (text $\rightarrow$ tokenizer). This \emph{cross-path inconsistency} (Figure~\ref{fig:motivation}) leads to substantial performance degradation: on VTCBench~\cite{zhao2025vtcbenchvisionlanguagemodelsunderstand}, Qwen3-VL-8B under VTC scores only \textbf{35.10}, far below the same backbone's native text-input performance of \textbf{55.60}. We argue that this reveals the true bottleneck of VTC: the core challenge is not ``how to compress more,'' but \textbf{how to align the visual encoder's representations with those of native text}.

Crucially, since both paths share the same MLLM backbone, the model's strong text-path behavior can naturally supervise its image-path behavior, requiring no external teachers or additional annotations. Based on this insight, we propose \textbf{SPIRAL} (\textbf{S}elf-improving \textbf{P}ath \textbf{I}ntegration and \textbf{R}ealignment), a self-supervised alignment framework operating at two complementary granularities. At the \textbf{token level}, on-policy distillation (OPD)~\cite{agarwal2024onpolicydistillationlanguagemodels} progressively aligns the image path's output distributions to those of the text path along self-generated trajectories, providing fine-grained, step-by-step correction. At the \textbf{sequence level}, DPO-based preference optimization~\cite{rafailov2023direct} treats text-path responses as preferred and image-path responses as dispreferred, shaping overall output quality with a global training signal. The two granularities are deliberately complementary: token-level supervision excels at precise retrieval tasks where local accuracy matters, while sequence-level supervision scales better and captures holistic reasoning and memory patterns.

Experiments on retrieval, reasoning, and memory benchmarks~\cite{zhao2025vtcbenchvisionlanguagemodelsunderstand,hsieh2024rulerwhatsrealcontext,bai2024longbenchbilingualmultitaskbenchmark} validate this approach. SPIRAL raises the overall VTCBench score from \textbf{35.10} to \textbf{54.02}, approaching the native text-input performance (\textbf{55.60}) and outperforming all existing VTC methods including models up to 30$\times$ larger. 

Our contributions are as follows:
\begin{itemize}[left=0.2em]
    \setlength{\itemsep}{0.05em}
    \setlength{\parskip}{0.1em}
    \item We identify \emph{cross-path inconsistency}, the representational gap between rendered-image and native-text inputs, as a critical yet overlooked bottleneck of VTC, reframing the core challenge from ``how to compress better'' to ``how to align the visual encoder with native text semantics.''
    \item We propose SPIRAL, a self-supervised alignment framework that uses the model's own text-path behavior to supervise its image-path behavior, combining token-level on-policy distillation with sequence-level preference optimization.
    \item SPIRAL closes most of the cross-path gap ($35.10 \rightarrow 54.02$ vs.\ text-mode $55.60$), outperforming all existing VTC methods, and generalizes to out-of-domain benchmarks.
\end{itemize}
\section{Related Work}

\subsection{Long-Context Efficiency and Compression}

Long-context efficiency has been pursued through several complementary directions, including sparse or linear attention~\cite{beltagy2020longformerlongdocumenttransformer,zaheer2020big,kitaev2020reformer,choromanski2022rethinkingattentionperformers,ding2023longnetscalingtransformers1000000000}, optimized exact attention~\cite{dao2022flashattention}, position extrapolation~\cite{press2021trainshort,chen2023extendingcontextwindowlarge,peng2023yarn,ding2024longropeextendingllmcontext}, retrieval augmentation~\cite{lewis2020retrieval,borgeaud2022improvinglanguagemodelsretrieving,khandelwal2019generalization}, and memory mechanisms~\cite{wu2022memorizingtransformers,rae2019compressivetransformerslongrangesequence,bulatov2022recurrentmemorytransformer,packer2024memgptllmsoperatingsystems}. Related context-reduction strategies such as LLMLingua~\cite{jiang2023llmlingua} and CEPE~\cite{yen2024longcontextlanguagemodelingparallel} further reduce effective input cost.

VTC offers a different trade-off: instead of compressing within the text modality, it renders long text into high-density images and processes them with a vision encoder~\cite{wei2025deepseekocrcontextsopticalcompression,cheng2025glyphscalingcontextwindows,li2025textorpixels}. Representative VTC systems include VisInContext~\cite{wang2024visincontext}, SEEKER~\cite{lu2024textpixeladvancinglongcontext}, VIST~\cite{xing2025visioncentrictokencompressionlarge}, VIST2~\cite{jiao2026globalcontextcompressioninterleaved}, and VTC-R1~\cite{wang2026vtcr1visiontextcompressionefficient}. Benchmarks such as VTCBench~\cite{zhao2025vtcbenchvisionlanguagemodelsunderstand} and ZeroSense~\cite{gao2026zerosensehowvisionmatterslong} further show that current VLMs still struggle with long-range understanding under rendered inputs, and that OCR accuracy does not directly translate to downstream performance. Our work addresses a complementary question: how to train VLMs so that behavior on rendered-image inputs remains faithful to behavior on native text inputs.

\subsection{Distillation and Cross-Modal Alignment}

Knowledge distillation transfers capability from a teacher to a student.  On-policy distillation~\cite{agarwal2024onpolicydistillationlanguagemodels} is particularly relevant because it aligns on self-generated trajectories and reduces the train--test mismatch of conventional offline distillation, which directly motivates our OPD design.

Cross-modal alignment in VLMs typically relies on contrastive pretraining~\cite{radford2021learning,jia2021scalingvisualvisionlanguagerepresentation}, bridging modules between frozen vision encoders and language models~\cite{li2023blip2,alayrac2022flamingo}, or cross-modal self-distillation~\cite{feng2024alignkd,kim2024cosmos,naeem2024silc}. However, these settings usually involve asymmetry across model scale, architecture, or modality content. In VTC, the same backbone receives semantically identical content through two input paths (native text and rendered image), making the problem one of cross-path alignment rather than conventional cross-model alignment. This setting is also supported by recent advances in document understanding and text-rich image modeling~\cite{kim2021ocrfree,lee2022pix2struct,huang2022layoutlmv3,li2023trocr,wei2024generalocrtheoryocr20}, which show that vision models can effectively parse dense rendered text.

\subsection{Preference Optimization for Multimodal Models}

Preference optimization has emerged as a practical alternative to RLHF. DPO~\cite{rafailov2023direct} formulates alignment as classification over preferred and dispreferred responses, while ORPO, SimPO, and KTO simplify or generalize the objective~\cite{hong2024orpomonolithicpreferenceoptimization,meng2024simpo,ethayarajh2024ktomodelalignmentprospect}.

In multimodal models, a central challenge is modality imbalance. mDPO~\cite{wang2024mdpoconditionalpreferenceoptimization} shows that models may ignore visual input and rely mainly on text signals, MFPO~\cite{jiang2025modalityfairpreferenceoptimizationtrustworthy} encourages fairer modality contributions, and CHiP~\cite{fu2025chipcrossmodalhierarchicaldirect} extends preference optimization across multiple granularities. In our VTC setting, preference pairs arise naturally from the two input paths: outputs from native-text inputs serve as the preferred target ($y^+$), while outputs from rendered-image inputs serve as the dispreferred target ($y^-$). This self-generated signal requires no human annotation or external reward model. \textcolor{black}{Moreover, unlike typical multimodal preference settings where the visual and textual content differ, our preference pairs compare two responses to semantically identical content delivered through different modalities, making the preference signal inherently clean and well-defined.} By comparing token-level OPD with sequence-level DPO under a shared framework, we systematically study alignment granularity in VTC.

\begin{figure*}[t]
  \centering
  \includegraphics[width=\textwidth]{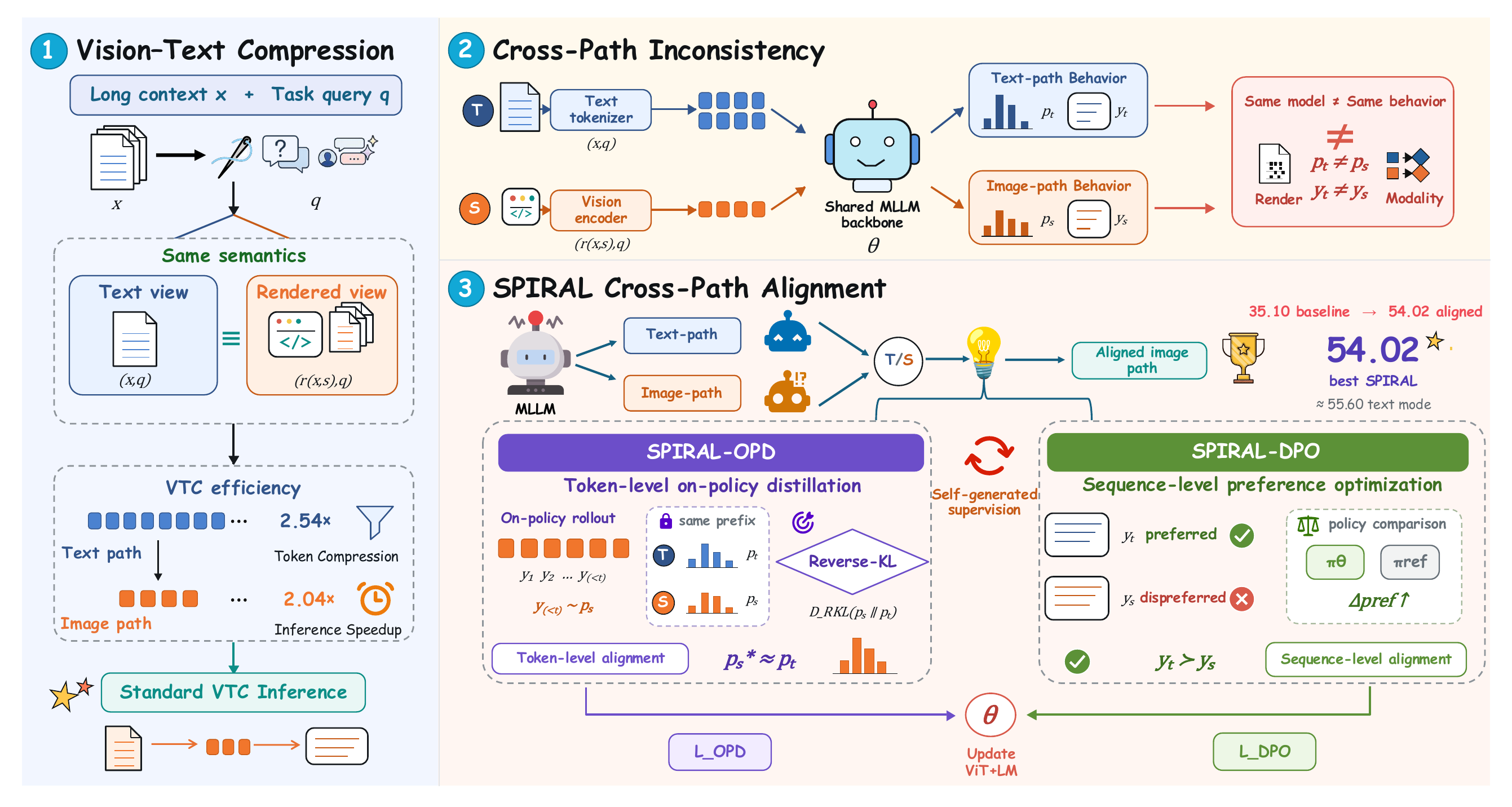}
  \caption{Overview of the SPIRAL framework. The same MLLM backbone processes both native text input and rendered-image input. SPIRAL aligns the two at token level via on-policy distillation (OPD) and at sequence level via preference optimization (DPO). Both mechanisms derive supervision entirely from the model's own text-input behavior.}
  \Description{Overview of the SPIRAL framework showing rendering, dual-input architecture, OPD token-level alignment, and DPO sequence-level alignment.}
  \label{fig:framework}
\end{figure*}

\section{Method}

\subsection{Preliminaries}

Vision-Text Compression (VTC) renders a long textual context $x$ into a high-density image $r(x,s)$ via an HTML/CSS rendering engine, where $s$ denotes rendering parameters such as font size, page dimensions, and line spacing (detailed in Section~\ref{sec:setup}). The rendered image is then encoded by the vision encoder (ViT) of a multimodal large language model (MLLM), which maps it to a sequence of visual tokens. Because a rendered page can pack thousands of characters into a single image, the resulting visual token sequence is substantially shorter than the native text token sequence for the same content, achieving significant token compression.

Because the MLLM already contains both a vision encoder and a text tokenizer, the same backbone can process the same content $x$ through two input paths. Given a task query $q$ and a response $y = (y_1, \ldots, y_T)$, the two paths yield:
\begin{itemize}
    \item \emph{Text path}: the model receives $(x, q)$ as native text and produces next-token distributions $p_{\mathcal{T}}(y_t \mid y_{<t}, x, q)$.
    \item \emph{Image path}: the model receives $(r(x,s), q)$ with the context rendered as an image and produces $p_{\mathcal{S}}(y_t \mid y_{<t}, r(x,s), q)$.
\end{itemize}

\noindent Since both paths share the same backbone and $x$ and $r(x,s)$ carry identical semantics, an ideal VTC system should satisfy $p_{\mathcal{T}}(\cdot) \approx p_{\mathcal{S}}(\cdot)$. We term the deviation between the two distributions \emph{cross-path inconsistency} (Figure~\ref{fig:motivation}).

\subsection{Motivation}

In practice, we observe that the image path produces substantially worse outputs than the text path on the same content. A natural question arises: is this cross-path gap caused by the ViT encoder capturing visual attributes rather than linguistic semantics, or simply by information loss during rendering? The answer has direct implications for how to close the gap.

If the gap stems from the ViT encoder being sensitive to visual appearance rather than from information loss during rendering, then changing the \emph{rendering style} while keeping the text \emph{identical} should degrade the image path's performance. We test this hypothesis with a rendering perturbation study. We apply three levels of perturbation to the rendering parameters $s$ while keeping the textual content $x$ fixed:
\begin{itemize}[left=0.2em,itemsep=0.1em]
    \item \textbf{L1} (mild): font size $\pm$1\,px, minor color and spacing changes.
    \item \textbf{L2} (moderate): font size $\pm$3\,px, font family swap, noticeable spacing and color shifts.
    \item \textbf{L3} (severe): font size $\pm$5\,px, aggressive font, color, and spacing changes.
\end{itemize}

\begin{table}[t]
\centering
\small
\renewcommand{\arraystretch}{1.2}
\caption{Rendering perturbation on the unaligned baseline at 1k--4k context lengths, with identical text rendered under increasing visual perturbation (L1--L3).}
\label{tab:perturbation}
\begin{tabular*}{\columnwidth}{l @{\extracolsep{\fill}} rrrr r}
\toprule
& \multicolumn{4}{c}{\textbf{Perturbation Level}} & \\
\cmidrule(lr){2-5}
\textbf{Task}
& \textbf{None}
& \textbf{L1}
& \textbf{L2}
& \textbf{L3}
& $\boldsymbol{\Delta}$ \\
\midrule
Retrieval
& 83.71
& 81.33
& 73.26
& 60.05
& \textcolor{red}{--23.7} \\[1pt]

Reasoning
& 10.20
& 10.08
& 11.60
& 11.06
& \textcolor{gray}{+0.9} \\[1pt]

Memory
& 24.87
& 21.88
& 19.75
& 18.58
& \textcolor{red}{--6.3} \\

\cmidrule(lr){1-6}
\rowcolor{lightgray}
\textbf{Total}
& \textbf{39.59}
& 37.76
& 34.87
& \textbf{29.90}
& \textcolor{red}{\textbf{--9.7}} \\
\bottomrule
\end{tabular*}
\end{table}

Table~\ref{tab:perturbation} confirms the hypothesis: the unaligned model is \emph{extremely fragile} to rendering perturbations. The total score drops from 39.59 to 29.90 under L3 (a 24.5\% relative decline), even though the underlying text is unchanged. Retrieval suffers the most, falling from 83.71 to 60.05, indicating that the model's ability to locate information depends heavily on specific visual patterns rather than linguistic content. This establishes that the cross-path gap arises because the ViT encoder is sensitive to \emph{rendering style} (font, spacing, layout) rather than to the \emph{linguistic content} carried by the rendered text.

This diagnosis directly suggests a remedy. Since both paths share the same backbone, and the text path already produces strong outputs on the same content, the model's own text-path behavior provides a natural supervision signal for the image path, without requiring any external teacher, additional annotations, or reward model. Based on this insight, we propose \textbf{SPIRAL} (\textbf{S}elf-improving \textbf{P}ath \textbf{I}ntegration and \textbf{R}ealignment), which explicitly aligns $p_{\mathcal{S}}$ toward $p_{\mathcal{T}}$ (Figure~\ref{fig:framework}). We design alignment at two complementary granularities, detailed in the following subsections.

\subsection{Token-Level Alignment}

The first granularity targets per-step distributional alignment. A straightforward approach would be standard knowledge distillation on pre-collected text-path targets; however, such off-policy targets may not reflect the model's actual behavior under image-path input. On-policy distillation (OPD)~\cite{agarwal2024onpolicydistillationlanguagemodels} addresses this by performing alignment on trajectories that the model itself generates from the image path, ensuring supervision targets the states most relevant to inference.

Concretely, we first sample a response from the rendered-image input:
\begin{equation}
    y \sim p_{\mathcal{S}}(\cdot \mid r(x,s), q).
\end{equation}
We then evaluate the native-text distributions at the same decoding prefixes. The distillation loss is:
\begin{equation}
    \mathcal{L}_{\mathrm{OPD}} =
    \sum_{t \in \mathcal{Y}}
    D
    \Big(
    p_{\mathcal{T}}(\cdot \mid y_{<t}, x, q),\,
    p_{\mathcal{S}}(\cdot \mid y_{<t}, r(x,s), q)
    \Big),
\end{equation}
where $\mathcal{Y}$ denotes the response-token positions and $D(\cdot,\cdot)$ is a token-level divergence. Note that $D(p_{\mathcal{T}}, p_{\mathcal{S}})$ denotes $\mathrm{KL}(p_{\mathcal{S}} \| p_{\mathcal{T}})$ (reverse KL), which is mode-seeking: it encourages $p_{\mathcal{S}}$ to concentrate on the high-probability regions of $p_{\mathcal{T}}$ rather than spreading mass across the entire support. This is well-suited to cross-path alignment, where the rendered-image distribution is typically more dispersed than the native-text distribution and needs to be sharpened toward the correct modes. We also experiment with forward KL and Jensen--Shannon divergence in our ablations (Table~\ref{tab:ablation-opd}).

As the rendered-image distributions improve through training, the model generates higher-quality trajectories, which in turn provide more informative alignment targets. This creates a virtuous cycle: better alignment leads to better trajectories, which in turn yield sharper supervision targets. OPD provides dense, per-step supervision, making it particularly suited for tasks requiring fine-grained faithfulness such as retrieval from long rendered contexts. However, on-policy sampling requires generating a full response for each training example before computing the loss, making OPD substantially more expensive per iteration than off-policy methods.

\subsection{Sequence-Level Alignment}

The second granularity targets holistic response quality. While token-level OPD aligns distributions at each step, locally reasonable token choices can still compound into globally incoherent responses. We therefore complement OPD with sequence-level alignment via Direct Preference Optimization (DPO)~\cite{rafailov2023direct}.

For a given rendered-image input $(r(x,s), q)$, we construct preference pairs from the model's own outputs: the native-text response $y^{+}$ (preferred) is generated by feeding the native text $x$ to the model, while the rendered-image response $y^{-}$ (dispreferred) is generated from the rendered image $r(x,s)$. Both responses are sampled from the same model checkpoint using temperature sampling. To ensure meaningful preference signals, we filter out pairs where the two responses are nearly identical or where the native-text response is clearly degenerate. Note that although $y^{+}$ is generated from native-text input, the DPO objective conditions both $y^{+}$ and $y^{-}$ on the rendered-image input $r(x,s)$: the goal is to shift the \emph{image-input policy} toward producing text-quality responses. The policy $\pi_{\theta}$ is then optimized relative to a frozen reference $\pi_{\mathrm{ref}}$:
\begin{equation}
    \Delta_{\theta}
    =
    \log \pi_{\theta}(y^{+}\mid r(x,s), q)
    -
    \log \pi_{\theta}(y^{-}\mid r(x,s), q),
\end{equation}
\begin{equation}
    \Delta_{\mathrm{ref}}
    =
    \log \pi_{\mathrm{ref}}(y^{+}\mid r(x,s), q)
    -
    \log \pi_{\mathrm{ref}}(y^{-}\mid r(x,s), q),
\end{equation}
\begin{equation}
    \mathcal{L}_{\mathrm{DPO}}
    =
    -\log \sigma \big( \beta (\Delta_{\theta} - \Delta_{\mathrm{ref}}) \big),
\end{equation}
where $\beta$ controls the sharpness of the preference margin. Unlike OPD, DPO provides a coarser but more holistic signal: it does not force exact distribution matching but steers the rendered-image policy toward text-consistent responses at the sequence level, making it particularly effective for reasoning and memory tasks where overall coherence matters. Because DPO does not require on-policy sampling, it is computationally cheaper per iteration and can scale to larger training sets more easily than OPD.

\subsection{Training and Inference}

\noindent\textbf{Training.} All alignment methods share the same dual-view training data: each sample pairs a rendered-image input (student view) with the corresponding native-text input (teacher view). The three methods differ only in how they utilize the teacher's signal. As a baseline, standard supervised fine-tuning (SFT) directly trains the model to produce the teacher-view response given the student-view input, the simplest off-policy form of cross-path alignment. OPD and DPO go further by incorporating on-policy and preference-based signals, respectively. We study OPD ($\mathcal{L}_{\mathrm{OPD}}$) and DPO ($\mathcal{L}_{\mathrm{DPO}}$) independently to isolate the contribution of each alignment granularity.

\noindent\textbf{Inference.} Both mechanisms operate \emph{only during training}. At inference, the aligned model runs identically to the unaligned baseline, with no additional forward passes, no reference model, and no teacher queries, preserving the full efficiency benefit of VTC.


\begin{table*}[t]
\centering
\renewcommand{\arraystretch}{1.2}
\setlength{\tabcolsep}{16pt}
\caption{
Comparison with existing models on VTCBench under predefined rendering.
SPIRAL denotes our cross-path alignment framework (Section~3).
Shaded rows denote SPIRAL-aligned models.
\first{Bold} and \underline{underlined} scores indicate the best and second-best SPIRAL results for each backbone, respectively.}
\label{tab:model-comparison}
\begin{tabular}{l l c c c c}
\toprule
\textbf{Model} & \textbf{Type} & \textbf{Retrieval} & \textbf{Reasoning} & \textbf{Memory} & \textbf{Total} \\
\midrule
\textcolor{gray}{\emph{Qwen3-VL-8B (text mode)}} & \textcolor{gray}{\emph{---}} & \textcolor{gray}{\emph{99.17}} & \textcolor{gray}{\emph{37.26}} & \textcolor{gray}{\emph{30.36}} & \textcolor{gray}{\emph{55.60}} \\
\midrule

Kimi-VL-A3B & open & 54.19 & 9.97 & 32.56 & 32.24 \\
Gemma3-27B & open & 18.22 & 2.03 & 13.12 & 11.12 \\
GLM-4.1V-9B-Thinking & open & 8.56 & 5.49 & 1.39 & 5.15 \\
Glyph & open & 82.07 & 17.62 & 2.59 & 34.10 \\
InternVL3.5-38B & open & 20.44 & 4.82 & 20.01 & 15.09 \\
Qwen2.5-VL-7B & open & 76.37 & 11.90 & 31.72 & 40.00 \\
Qwen2.5-VL-72B & open & 84.43 & 28.63 & 31.72 & 48.26 \\
Qwen3-VL-235B-A22B & open & 78.72 & 7.30 & 30.85 & 38.96 \\

\cmidrule(lr){1-6}

Gemini-2.5-Pro & proprietary & 56.99 & 41.18 & 25.42 & 41.20 \\
GPT-5 & proprietary & 33.94 & 14.04 & 29.29 & 25.76 \\

\cmidrule(lr){1-6}

Qwen3-VL-8B (baseline) & open & 74.50 & 5.95 & 24.86 & 35.10 \\
\rowcolor{lightgray}
\textbf{+ SPIRAL-DPO (Qwen3-VL-8B)} & open & \underline{83.01} & \textbf{44.17} & \textbf{34.14} & \underline{53.77} \\
\rowcolor{lightgray}
\textbf{+ SPIRAL-OPD (Qwen3-VL-8B)} & open & \textbf{89.34} & \underline{40.74} & \underline{31.98} & \textbf{54.02} \\

InternVL3.5-8B (baseline) & open & 13.83 & 3.15 & 23.40 & 13.46 \\
\rowcolor{lightgray}
\textbf{+ SPIRAL-DPO (InternVL3.5-8B)} & open & \textbf{24.99} & \underline{8.48} & \underline{27.53} & \underline{20.33} \\
\rowcolor{lightgray}
\textbf{+ SPIRAL-OPD (InternVL3.5-8B)} & open & \underline{23.79} & \textbf{13.52} & \textbf{30.52} & \textbf{22.61} \\

\bottomrule
\end{tabular}
\end{table*}

\section{Experiment}

\subsection{Experimental Setup}
\label{sec:setup}

\noindent\textbf{Backbone.} We use Qwen3-VL-8B as the primary MLLM backbone for our main experiments. The same model is used for both native-text input and rendered-image input; only the input modality changes. As a native-text reference, we also report the performance of the same backbone when it processes raw text directly without rendering (Qwen3-VL-8B in text mode). To assess cross-backbone generality, we additionally train both SPIRAL variants on InternVL3.5-8B under the same rendering and paired-view alignment protocol.

\noindent\textbf{Rendering.} Long textual contexts are rendered into page images with an HTML/CSS engine (Playwright/Chromium). Each page is rasterized at $896\times896$ pixels (96\,DPI) and stored as JPEG (quality~85). We use Helvetica 12\,px with a line height of 1.2, which yields approximately $2.55\times$ token compression relative to native text tokenization. Only the long context is rendered and the task query itself remains in native text.

\begin{table}[t]
\centering
\renewcommand{\arraystretch}{1.2}
\setlength{\tabcolsep}{5pt}
\caption{Results on generalization benchmarks. \first{Bold}: best VTC method. \second{Underline}: second best.}
\label{tab:main-results}
\begin{tabular}{l c c c c}
\toprule
\textbf{Method} & \textbf{LBench} & \textbf{TQA} & \textbf{2Wiki} & \textbf{GSM8K} \\
\midrule
\textcolor{gray}{\emph{Text mode}} & \textcolor{gray}{\emph{51.78}} & \textcolor{gray}{\emph{92.16}} & \textcolor{gray}{\emph{56.17}} & \textcolor{gray}{\emph{94.80}} \\[2pt]
Baseline & 34.62 & 85.11 & 40.94 & 88.70 \\
\midrule
SFT & 36.50 & 88.20 & 36.31 & 89.10 \\
\rowcolor{lightgray}
OPD & \second{37.07} & \first{88.88} & \second{47.14} & \second{89.60} \\
\rowcolor{lightgray}
DPO & \first{37.98} & \second{88.56} & \first{48.16} & \first{90.20} \\
\bottomrule
\end{tabular}
\end{table}

\noindent\textbf{Efficiency.} 
Under our rendering setup, VTC reduces the average token count from 20,673 to 8,145 ($2.54\times$ compression), corresponding to a $2.04\times$ inference speedup in forward-pass time. 
Since SPIRAL is applied only during training, aligned models incur the same inference cost as the unaligned VTC baseline.

\begin{figure*}[t]
  \centering
  \includegraphics[width=\textwidth]{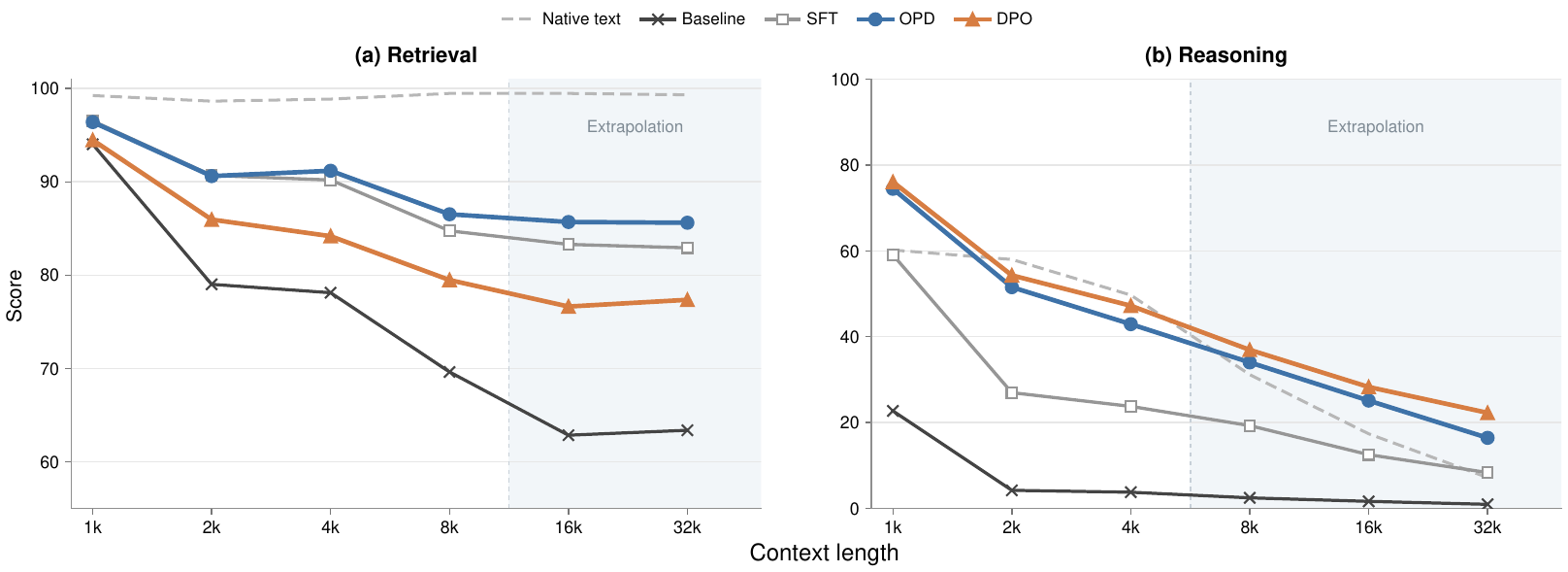}
  \caption{Performance vs.\ context length (1k--32k) for retrieval (left) and reasoning (right). OPD maintains strong retrieval at all lengths; DPO's reasoning advantage grows with context length, surpassing even the native text-input score at 32k.}
  \Description{Line plots showing retrieval and reasoning performance vs context length for all methods.}
  \label{fig:context-length}
\end{figure*}

\noindent\textbf{Training data and alignment setting.}
We construct 115k training samples that span three task types: \emph{Retrieval} (40k, PG19 haystack + RULER needles at context lengths of 1k--8k), \emph{Reasoning} (21k, PG19 haystack + NoLiMa needles at 1k--4k), and \emph{Memory} (54k, synthetic LoCoMo-style conversations generated by Gemini 2.5 Pro). Each sample provides two views of the same instance: a \textbf{student view}, where the long context is rendered into images (\texttt{<image>} tokens + text query), and a \textbf{teacher view}, where the same context is given as native text.

SPIRAL uses the model's native-text behavior to supervise the rendered-image path through the shared backbone: SPIRAL-OPD performs token-level on-policy distillation along student-generated trajectories, whereas SPIRAL-DPO applies sequence-level preference optimization between teacher- and student-view responses. SFT serves as an off-policy baseline; all three methods share the same dual-view instances and text-derived supervision, with disjoint needle vocabularies and incorrectly answered teacher samples filtered out.

\noindent\textbf{Training details.} 
For OPD, we use reverse KL as the default divergence. For DPO, we set $\beta=0.1$. 
Both SPIRAL variants fine-tune the full model (ViT + LM) with AdamW. 
We evaluate at training sizes of 10k, 30k, and 100k examples. 
OPD is not evaluated at 100k because on-policy trajectory sampling is more expensive at scale.

\noindent\textbf{Evaluation.} 
We evaluate on three groups from VTCBench~\cite{zhao2025vtcbenchvisionlanguagemodelsunderstand}, each targeting a distinct aspect of long-context understanding under VTC:
\begin{itemize}[
    left=0.2em,
    topsep=0pt,
    partopsep=0pt,
    parsep=0pt,
    itemsep=0.1em
]
    \item \emph{Retrieval} (RULER~\cite{hsieh2024rulerwhatsrealcontext}): locating specific information injected into long rendered contexts (needle-in-a-haystack), testing precise extraction ability.
    \item \emph{Reasoning} (NoLiMa): answering questions that require multi-step inference over long rendered book passages, testing global comprehension.
    \item \emph{Memory} (LoCoMo): recalling details from long rendered conversation histories across sub-tasks (MultiHop, Temporal, OpenDomain, SingleHop), testing long-range information retention.
\end{itemize}

\noindent Context lengths range from \textbf{1k} to \textbf{32k} tokens. We additionally evaluate on LongBench~\cite{bai2024longbenchbilingualmultitaskbenchmark} to test generalization. We report per-group scores and an overall \emph{Total} score.

\subsection{Main Results}

Table~\ref{tab:model-comparison} compares our aligned models with existing systems on VTCBench under a common predefined rendering setup. 
Most models, including strong proprietary systems, perform poorly under VTC, suggesting that long-context understanding with rendered inputs remains challenging across model families. 
Under this fixed setting, both SPIRAL variants substantially improve Qwen3-VL-8B over the unaligned baseline. 
SPIRAL-OPD achieves the best overall score among VTC methods (54.02), while SPIRAL-DPO yields the strongest reasoning and memory scores (44.17 and 34.14), bringing performance close to the native-text reference (55.60).
To further evaluate cross-backbone generality, we also apply both SPIRAL variants to InternVL3.5-8B. As shown in Table~\ref{tab:model-comparison}, SPIRAL-OPD and SPIRAL-DPO raise its total score from 13.46 to 22.61 and 20.33, respectively, with consistent improvements across all three task groups.
These gains provide preliminary evidence that the cross-path alignment signal is not specific to Qwen3-VL-8B.

Table~\ref{tab:main-results} reports out-of-domain generalization of each method. 
On all four benchmarks, the unaligned baseline trails the native-text reference, and both SPIRAL variants recover a substantial portion of this gap. 
DPO performs best on LongBench~\cite{bai2024longbenchbilingualmultitaskbenchmark} (37.98), 2WikiMultihopQA~\cite{ho-etal-2020-constructing} (48.16), and GSM8K~\cite{cobbe2021trainingverifierssolvemath} (90.20), while OPD performs best on TriviaQA~\cite{joshi2017triviaqalargescaledistantly} (88.88). 
SFT improves over the baseline on three benchmarks but drops below it on 2WikiMultihopQA (36.31 vs.\ 40.94), indicating that simple off-policy alignment can be brittle on tasks requiring multi-hop reasoning. 
Overall, the results suggest that explicit cross-path alignment improves both in-domain VTCBench performance and out-of-domain generalization.

\subsection{Scaling Behavior}

\begin{figure}[t]
  \centering
  \includegraphics[width=\columnwidth]{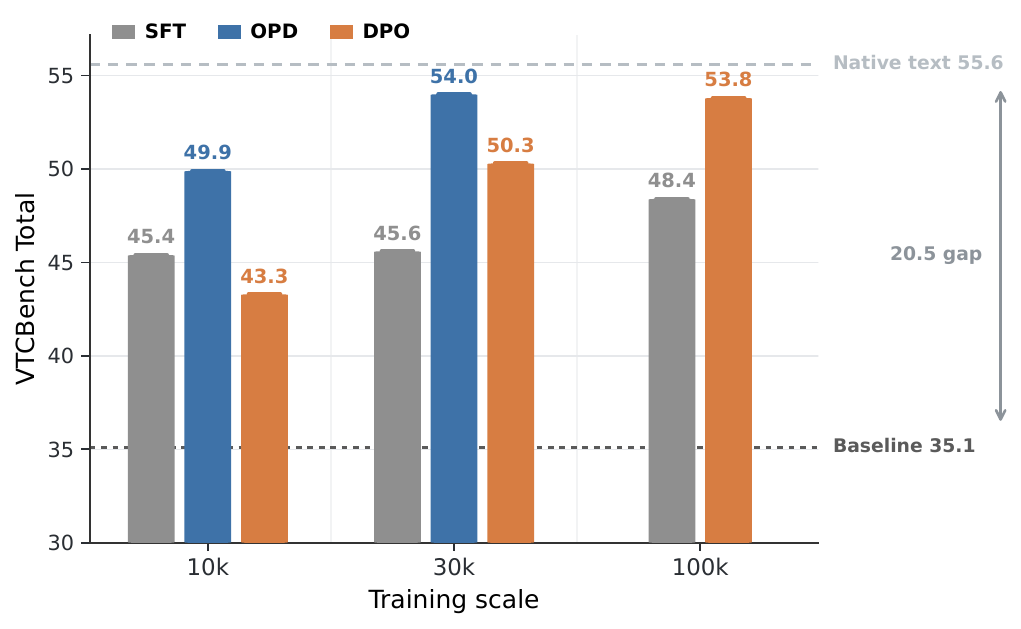}
  \caption{Effect of training scale on overall performance. OPD converges quickly (sample-efficient); DPO scales steadily with more data.}
  \Description{Line plot showing total score vs training data size for OPD, SFT, and DPO methods.}
  \label{fig:scaling}
\end{figure}

Figure~\ref{fig:scaling} shows how the three methods scale with training data. OPD achieves strong performance with only 10k examples (49.94) and reaches 54.02 at 30k, demonstrating high sample efficiency. Due to the substantially higher computational cost of on-policy trajectory sampling, we do not evaluate OPD at 100k; however, the consistent improvement from 10k to 30k suggests that OPD has not yet saturated and may benefit from further scaling. DPO starts lower but scales steadily: 43.29 at 10k, 50.27 at 30k, and 53.77 at 100k. SFT benefits modestly from additional data ($45.37 \rightarrow 48.43$). This pattern reflects the nature of each alignment signal: token-level OPD provides dense per-step supervision enabling rapid convergence at small scale, while sequence-level DPO captures global patterns that improve with more contrastive pairs.
This difference explains OPD's stronger sample efficiency at smaller scales and DPO's continued gains with more preference pairs.
\textcolor{black}{From a practical standpoint, OPD is attractive when the number of training examples is limited and on-policy rollout compute is affordable, whereas DPO is computationally cheaper per iteration and scales more readily to larger datasets.}

\subsection{Context Length Generalization}

Figure~\ref{fig:context-length} shows performance across context lengths from 1k to 32k. In retrieval, the baseline degrades sharply from 94.02 at 1k to 63.37 at 32k. OPD maintains strong retrieval across all lengths, achieving 85.61 even at 32k, a 22-point improvement over the baseline.

The most striking pattern emerges in reasoning. The baseline collapses from 22.70 at 1k to 0.94 at 32k, indicating that unaligned VTC completely fails for long-context reasoning. DPO shows the strongest recovery, reaching 22.26 at 32k, surpassing even the text-mode score at that length (7.18). DPO's advantage over OPD widens with context length: the two methods are close at 1k (76.05 vs.\ 74.45), but by 32k DPO opens a clear margin (22.26 vs.\ 16.43). This confirms that sequence-level alignment becomes increasingly important as contexts grow longer.

Notably, the training data contains contexts only up to 8k tokens (retrieval) and 4k tokens (reasoning), yet alignment benefits generalize to 16k and 32k, lengths never seen during training. This suggests that SPIRAL learns a general cross-path alignment capability rather than length-specific patterns. \textcolor{black}{We hypothesize that alignment teaches the ViT encoder to produce representations that are more semantically grounded regardless of input length, rather than memorizing length-specific correction patterns. This length generalization property is particularly valuable in practice, as it means SPIRAL can be trained on relatively short contexts while remaining effective on much longer documents at deployment.}

\begin{table}[t]
\centering
\renewcommand{\arraystretch}{1.3}
\caption{OPD ablation (10k). \first{Bold}: best. \second{Underline}: 2nd.}
\label{tab:ablation-opd}
\begin{tabular*}{\columnwidth}{l @{\extracolsep{\fill}} cccc}
\toprule
\textbf{Variant} & \textbf{Ret.} & \textbf{Rea.} & \textbf{Mem.} & \textbf{Total} \\
\midrule
\textcolor{gray}{\emph{Baseline}} & \textcolor{gray}{\emph{83.71}} & \textcolor{gray}{\emph{10.20}} & \textcolor{gray}{\emph{24.87}} & \textcolor{gray}{\emph{39.59}} \\
\midrule
\multicolumn{5}{l}{\emph{Parameter scope}} \\[2pt]
\rowcolor{lightgray}
~~Full (default) & \first{92.53} & \first{43.23} & \first{30.94} & \first{55.57} \\
~~ViT-only & 87.79 & 19.30 & 27.97 & 45.02 \\
~~LM-only & 91.11 & 36.01 & 29.60 & 52.24 \\[3pt]
\cmidrule(lr){1-5}
\multicolumn{5}{l}{\emph{Divergence function}} \\[2pt]
~~Reverse-KL (default) & \first{92.53} & \first{43.23} & \first{30.94} & \first{55.57} \\
~~Forward-KL & \second{92.05} & \second{42.80} & \second{30.73} & \second{55.19} \\
~~JSD(0.5) & 71.92 & 26.71 & 27.62 & 42.08 \\
\bottomrule
\end{tabular*}
\end{table}

\begin{table}[ht]
\centering
\renewcommand{\arraystretch}{1.1}
\caption{DPO ablation (30k). \first{Bold}: best. \second{Underline}: 2nd.}
\label{tab:ablation-dpo}
\begin{tabular*}{\columnwidth}{l @{\extracolsep{\fill}} cccc}
\toprule
\textbf{Variant} & \textbf{Ret.} & \textbf{Rea.} & \textbf{Mem.} & \textbf{Total} \\
\midrule
\textcolor{gray}{\emph{Baseline}} & \textcolor{gray}{\emph{83.71}} & \textcolor{gray}{\emph{10.20}} & \textcolor{gray}{\emph{24.87}} & \textcolor{gray}{\emph{39.59}} \\
\midrule
\multicolumn{5}{l}{\emph{Parameter scope}} \\[2pt]
\rowcolor{lightgray}
~~Full (default) & \second{89.23} & \first{54.80} & 27.51 & \first{57.18} \\
~~ViT-only & 85.02 & 25.41 & \first{31.20} & 47.21 \\
~~LM-only & \first{90.03} & \second{44.20} & 19.40 & \second{51.21} \\[3pt]
\cmidrule(lr){1-5}
\multicolumn{5}{l}{\emph{Preference filtering}} \\[2pt]
~~w/ filter (default) & 89.23 & 54.80 & 27.51 & \first{57.18} \\
~~w/o filter & 89.31 & 54.98 & \second{26.60} & 56.96 \\
\bottomrule
\end{tabular*}
\end{table}

\begin{figure*}[t]
  \centering
  \includegraphics[width=0.99\textwidth]{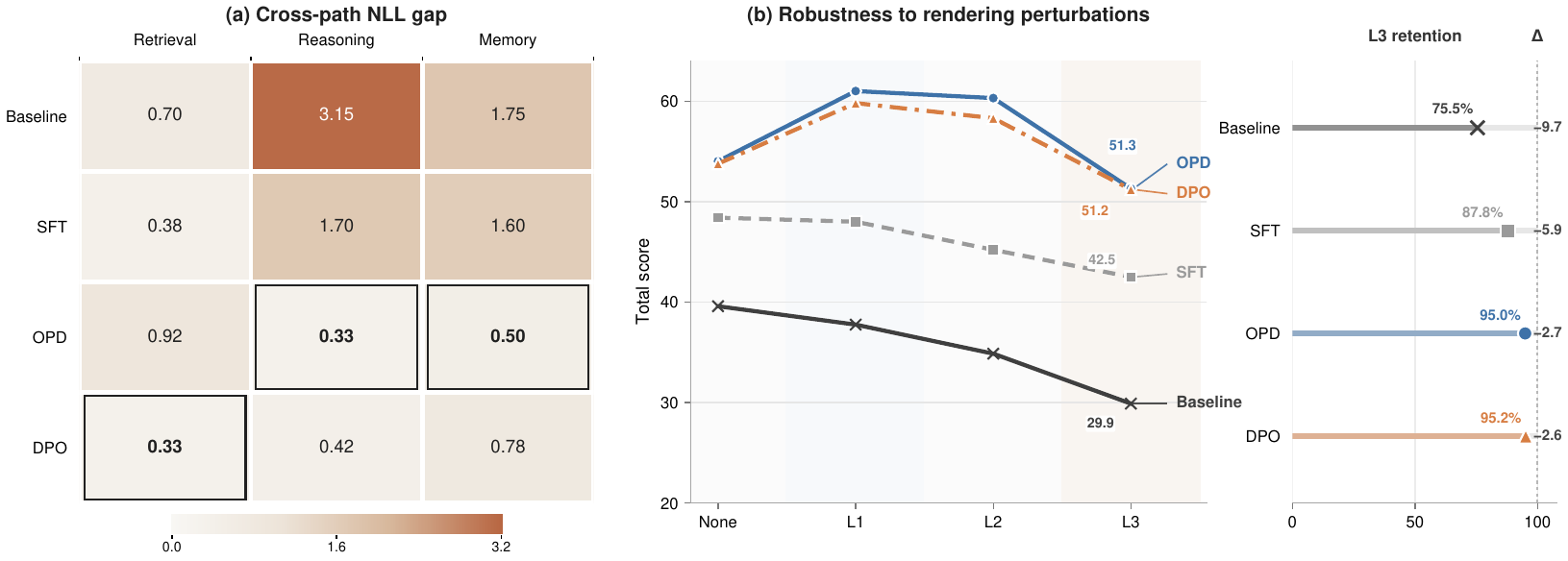}
  \caption{(a) Cross-path NLL gap across retrieval, reasoning, and memory (image NLL $-$ text NLL), where outlined cells indicate the best method per task. SPIRAL substantially narrows the gap, particularly on reasoning. (b) Perturbation robustness under None and L1--L3. The baseline degrades sharply, whereas SPIRAL preserves higher scores and substantially better L3 retention.}
  \Description{Two-panel figure: left shows a heatmap of cross-path NLL gaps across methods and tasks; right shows robustness curves under increasing rendering perturbations, together with L3 retention and score-drop summaries.}
  \label{fig:analysis}
\end{figure*}

\subsection{Ablation Studies}

We ablate key design choices for both OPD and DPO to understand which components drive their effectiveness. Specifically, we evaluate on reasoning and retrieval tasks with 1k, 2k, and 4k context lengths, as well as on the full memory tasks.

\noindent\textbf{OPD ablation (10k).} Table~\ref{tab:ablation-opd} varies the parameter scope and divergence function. Full-model fine-tuning (55.57) substantially outperforms ViT-only (45.02) and LM-only (52.24), confirming that both the vision encoder and language model must adapt for effective cross-path alignment. LM-only performs better than ViT-only, indicating the language model is the primary locus of misalignment. 
For divergence choice, reverse KL (55.57) and forward KL (55.19) perform comparably, while JSD(0.5) degrades severely (42.08), suggesting mode-seeking divergences are better suited for cross-path alignment. 

\noindent\textbf{DPO ablation (30k).} Table~\ref{tab:ablation-dpo} shows similar trends for parameter scope: full-model training (57.18) outperforms ViT-only (47.21) and LM-only (51.21). Interestingly, DPO-ViT-only achieves the best memory (31.20) but severely degrades reasoning (25.41), confirming full-model alignment is necessary for balanced performance. Removing preference pair quality filtering has minimal impact (57.18 vs.\ 56.96), suggesting DPO is robust to noisy pairs, likely because the native-text vs.\ rendered-image preference is inherently reliable.

\subsection{Analysis}

Beyond design-choice ablations, we examine whether SPIRAL achieves genuine distributional alignment and robustness to visual variation.

\noindent\textbf{Cross-path NLL gap.} Figure~\ref{fig:analysis}(a) quantifies the cross-path inconsistency via the NLL gap between text-input and image-input predictions. The unaligned baseline exhibits a large reasoning gap (3.15). SPIRAL substantially closes this gap: DPO reduces the average to 0.51, while SFT only reaches 1.24. This confirms that explicit alignment is more effective than off-policy adaptation at closing the distributional gap.

\noindent\textbf{Perturbation robustness.} Figure~\ref{fig:analysis}(b) extends the rendering perturbation study from Section~3.2 by evaluating aligned models. While the baseline drops 14.9 points under L3, OPD drops only 2.7 and DPO only 2.6. SFT drops 5.9, better than the baseline but substantially worse than SPIRAL. This confirms that cross-path alignment teaches the model to rely on linguistic semantics rather than visual patterns. \textcolor{black}{The near-identical robustness of OPD and DPO (2.7 vs.\ 2.6) suggests that both alignment granularities achieve a similar degree of semantic grounding, even though they arrive at it through different mechanisms. In contrast, SFT's intermediate robustness (5.9) indicates that off-policy supervision provides some regularization against visual artifacts but does not fully decouple rendering style from semantic content.}

\subsection{Summary}

Our experiments yield four findings. First, explicit cross-path alignment substantially outperforms off-policy SFT and closes most of the gap to native-text input. Second, on the primary backbone, OPD is more effective for retrieval and sample-efficient training, while DPO performs better on reasoning and memory and scales more steadily with data. Third, although trained on contexts of at most 8k tokens, SPIRAL generalizes to 32k. Finally, its gains transfer to out-of-domain benchmarks and substantially reduce the cross-path NLL gap, indicating genuine alignment rather than task-specific overfitting.

\section{Conclusion and Limitations}

\subsection{Conclusion}

We presented a cross-path alignment perspective on Vision-Text Compression, identifying \emph{cross-path inconsistency} as a key source of degradation when semantically identical content is processed through native-text and rendered-image paths. Based on this perspective, we proposed SPIRAL, a self-supervised framework that aligns the image path toward the text path using the model's own text-derived signals, without external teachers or additional annotations.

SPIRAL combines on-policy distillation (OPD) for token-level alignment and preference optimization (DPO) for sequence-level response quality. On VTCBench, it improves the overall score from \textbf{35.10} to \textbf{54.02}, approaching native text-input performance (\textbf{55.60}) and outperforming existing VTC methods. OPD is sample-efficient and excels at retrieval, while DPO scales more steadily and yields stronger reasoning and memory performance on the primary backbone. The improvements also transfer to out-of-domain benchmarks and InternVL3.5-8B, providing preliminary evidence of generalization beyond a single benchmark and backbone. Both mechanisms operate only during training, preserving the full inference efficiency of VTC. Overall, our results suggest that representational fidelity, rather than compression ratio alone, is central to effective vision-text compression.

\subsection{Limitations}

Our main experiments focus on Qwen3-VL-8B and a fixed rendering pipeline. Although results on InternVL3.5-8B provide preliminary cross-backbone evidence, broader validation across model scales, vision encoders, and rendering configurations remains necessary. We also do not isolate how rendering choices such as layout density, font size, and image resolution affect path consistency or the need for alignment. Finally, OPD is more sample-efficient but requires costly on-policy rollouts, whereas DPO scales more easily with larger datasets; the optimal trade-off between the two remains underexplored.

Future work should study: (1)~rendering--alignment interactions across diverse layouts and resolutions; (2)~hybrid OPD/DPO objectives; (3)~transfer to other cross-path settings; and (4)~iterative multi-round self-improvement.


\newpage
\bibliographystyle{ACM-Reference-Format}
\bibliography{references}

@misc{wei2025deepseekocrcontextsopticalcompression,
      title={DeepSeek-OCR: Contexts Optical Compression},
      author={Haoran Wei and Yaofeng Sun and Yukun Li},
      year={2025},
      eprint={2510.18234},
      archivePrefix={arXiv},
      primaryClass={cs.CV},
      url={https://arxiv.org/abs/2510.18234}, 
}

@misc{cheng2025glyphscalingcontextwindows,
      title={Glyph: Scaling Context Windows via Visual-Text Compression},
      author={Jiale Cheng and Yusen Liu and Xinyu Zhang and Yulin Fei and Wenyi Hong and Ruiliang Lyu and Weihan Wang and Zhe Su and Xiaotao Gu and Xiao Liu and Yushi Bai and Jie Tang and Hongning Wang and Minlie Huang},
      year={2025},
      eprint={2510.17800},
      archivePrefix={arXiv},
      primaryClass={cs.CV},
      url={https://arxiv.org/abs/2510.17800}, 
}

@misc{zhao2025vtcbenchvisionlanguagemodelsunderstand,
      title={VTCBench: Can Vision-Language Models Understand Long Context with Vision-Text Compression?},
      author={Hongbo Zhao and Meng Wang and Fei Zhu and Wenzhuo Liu and Bolin Ni and Fanhu Zeng and Gaofeng Meng and Zhaoxiang Zhang},
      year={2025},
      eprint={2512.15649},
      archivePrefix={arXiv},
      primaryClass={cs.CV},
      url={https://arxiv.org/abs/2512.15649}, 
}

@misc{gao2026zerosensehowvisionmatterslong,
      title={ZeroSense: How Vision Matters in Long Context Compression},
      author={Yonghan Gao and Zehong Chen and Lijian Xu and Jingzhi Chen and Jingwei Guan and Xingyu Zeng},
      year={2026},
      eprint={2603.11846},
      archivePrefix={arXiv},
      primaryClass={cs.CV},
      url={https://arxiv.org/abs/2603.11846}, 
}

@misc{wang2026vtcr1visiontextcompressionefficient,
      title={VTC-R1: Vision-Text Compression for Efficient Long-Context Reasoning},
      author={Yibo Wang and Yongcheng Jing and Shunyu Liu and Hao Guan and Rong-cheng Tu and Chengyu Wang and Jun Huang and Dacheng Tao},
      year={2026},
      eprint={2601.22069},
      archivePrefix={arXiv},
      primaryClass={cs.CL},
      url={https://arxiv.org/abs/2601.22069}, 
}

@misc{li2025textorpixels,
  title={Text or Pixels? It Takes Half: On the Token Efficiency of Visual Text Inputs in Multimodal LLMs},
  author={Yanhong Li and Zixuan Lan and Jiawei Zhou},
  year={2025},
  eprint={2510.18279},
  archivePrefix={arXiv},
  url={https://arxiv.org/abs/2510.18279}
}

@misc{lu2024textpixeladvancinglongcontext,
      title={From Text to Pixel: Advancing Long-Context Understanding in MLLMs},
      author={Yujie Lu and Xiujun Li and Tsu-Jui Fu and Miguel Eckstein and William Yang Wang},
      year={2024},
      eprint={2405.14213},
      archivePrefix={arXiv},
      primaryClass={cs.CV},
      url={https://arxiv.org/abs/2405.14213}, 
}

@article{wang2024visincontext,
      title={Leveraging Visual Tokens for Extended Text Contexts in Multi-Modal Learning},
      author={Wang, Alex Jinpeng and Li, Linjie and Lin, Yiqi and Li, Min  and Wang, Lijuan and Shou, Mike Zheng},
      journal={NeurIPS},
      year={2024}
}

@misc{xing2025visioncentrictokencompressionlarge,
      title={VIST: Vision-centric Token Compression in LLM},
      author={Ling Xing and Alex Jinpeng Wang and Rui Yan and Xiangbo Shu and Jinhui Tang},
      year={2025},
      eprint={2502.00791},
      archivePrefix={arXiv},
      primaryClass={cs.CL},
      url={https://arxiv.org/abs/2502.00791}, 
}

@misc{jiao2026globalcontextcompressioninterleaved,
      title={Global Context Compression with Interleaved Vision-Text Transformation},
      author={Dian Jiao and Jiaxin Duan and Shuai Zhao and Jiabing Leng and Yiran Zhang and Feng Huang},
      year={2026},
      eprint={2601.10378},
      archivePrefix={arXiv},
      primaryClass={cs.CV},
      url={https://arxiv.org/abs/2601.10378}, 
}

@article{dao2022flashattention,
  title={FlashAttention: Fast and Memory-Efficient Exact Attention with IO-Awareness},
  author={Dao, Tri and Fu, Dan and Ermon, Stefano and Rudra, Atri and R{\'e}, Christopher},
  journal={Advances in neural information processing systems},
  volume={35},
  pages={16344--16359},
  year={2022}
}

@misc{beltagy2020longformerlongdocumenttransformer,
      title={Longformer: The Long-Document Transformer},
      author={Iz Beltagy and Matthew E. Peters and Arman Cohan},
      year={2020},
      eprint={2004.05150},
      archivePrefix={arXiv},
      primaryClass={cs.CL},
      url={https://arxiv.org/abs/2004.05150}, 
}

@article{zaheer2020big,
  title={Big Bird: Transformers for Longer Sequences},
  author={Zaheer, Manzil and Guruganesh, Guru and Dubey, Kumar Avinava and Ainslie, Joshua and Alberti, Chris and Ontanon, Santiago and Pham, Philip and Ravula, Anirudh and Wang, Qifan and Yang, Li and others},
  journal={Advances in neural information processing systems},
  volume={33},
  pages={17283--17297},
  year={2020}
}

@article{kitaev2020reformer,
  title={Reformer: The Efficient Transformer},
  author={Kitaev, Nikita and Kaiser, {\L}ukasz and Levskaya, Anselm},
  journal={arXiv preprint arXiv:2001.04451},
  year={2020}
}

@misc{choromanski2022rethinkingattentionperformers,
      title={Rethinking Attention with Performers},
      author={Krzysztof Choromanski and Valerii Likhosherstov and David Dohan and Xingyou Song and Andreea Gane and Tamas Sarlos and Peter Hawkins and Jared Davis and Afroz Mohiuddin and Lukasz Kaiser and David Belanger and Lucy Colwell and Adrian Weller},
      year={2022},
      eprint={2009.14794},
      archivePrefix={arXiv},
      primaryClass={cs.LG},
      url={https://arxiv.org/abs/2009.14794}, 
}

@misc{ding2023longnetscalingtransformers1000000000,
      title={LongNet: Scaling Transformers to 1,000,000,000 Tokens},
      author={Jiayu Ding and Shuming Ma and Li Dong and Xingxing Zhang and Shaohan Huang and Wenhui Wang and Nanning Zheng and Furu Wei},
      year={2023},
      eprint={2307.02486},
      archivePrefix={arXiv},
      primaryClass={cs.CL},
      url={https://arxiv.org/abs/2307.02486}, 
}

@misc{press2021trainshort,
  title={Train Short, Test Long: Attention with Linear Biases Enables Input Length Extrapolation},
  author={Ofir Press and Noah A. Smith and Mike Lewis},
  year={2021},
  eprint={2108.12409},
  archivePrefix={arXiv},
  url={https://arxiv.org/abs/2108.12409}
}

@misc{chen2023extendingcontextwindowlarge,
      title={Extending Context Window of Large Language Models via Position Interpolation},
      author={Shouyuan Chen and Sherman Wong and Liangjian Chen and Yuandong Tian},
      year={2023},
      eprint={2306.15595},
      archivePrefix={arXiv},
      primaryClass={cs.CL},
      url={https://arxiv.org/abs/2306.15595}, 
}

@article{peng2023yarn,
  title={Yarn: Efficient context window extension of large language models},
  author={Peng, Bowen and Quesnelle, Jeffrey and Fan, Honglu and Shippole, Enrico},
  journal={arXiv preprint arXiv:2309.00071},
  year={2023}
}

@misc{ding2024longropeextendingllmcontext,
      title={LongRoPE: Extending LLM Context Window Beyond 2 Million Tokens},
      author={Yiran Ding and Li Lyna Zhang and Chengruidong Zhang and Yuanyuan Xu and Ning Shang and Jiahang Xu and Fan Yang and Mao Yang},
      year={2024},
      eprint={2402.13753},
      archivePrefix={arXiv},
      primaryClass={cs.CL},
      url={https://arxiv.org/abs/2402.13753}, 
}

@misc{lewis2020retrieval,
  title={Retrieval-Augmented Generation for Knowledge-Intensive NLP Tasks},
  author={Patrick Lewis and Ethan Perez and Aleksandra Piktus and Fabio Petroni and Vladimir Karpukhin and Naman Goyal and Heinrich Kuettler and Mike Lewis and Wen-tau Yih and Tim Rocktaeschel and Sebastian Riedel and Douwe Kiela},
  year={2020},
  eprint={2005.11401},
  archivePrefix={arXiv},
  url={https://arxiv.org/abs/2005.11401}
}

@misc{borgeaud2022improvinglanguagemodelsretrieving,
      title={Improving Language Models by Retrieving from Trillions of Tokens},
      author={Sebastian Borgeaud and Arthur Mensch and Jordan Hoffmann and Trevor Cai and Eliza Rutherford and Katie Millican and George van den Driessche and Jean-Baptiste Lespiau and Bogdan Damoc and Aidan Clark and Diego de Las Casas and Aurelia Guy and Jacob Menick and Roman Ring and Tom Hennigan and Saffron Huang and Loren Maggiore and Chris Jones and Albin Cassirer and Andy Brock and Michela Paganini and Geoffrey Irving and Oriol Vinyals and Simon Osindero and Karen Simonyan and Jack W. Rae and Erich Elsen and Laurent Sifre},
      year={2022},
      eprint={2112.04426},
      archivePrefix={arXiv},
      primaryClass={cs.CL},
      url={https://arxiv.org/abs/2112.04426}, 
}

@misc{khandelwal2019generalization,
  title={Generalization through Memorization: Nearest Neighbor Language Models},
  author={Urvashi Khandelwal and Omer Levy and Dan Jurafsky and Luke Zettlemoyer and Mike Lewis},
  year={2019},
  eprint={1911.00172},
  archivePrefix={arXiv},
  url={https://arxiv.org/abs/1911.00172}
}

@misc{wu2022memorizingtransformers,
      title={Memorizing Transformers},
      author={Yuhuai Wu and Markus N. Rabe and DeLesley Hutchins and Christian Szegedy},
      year={2022},
      eprint={2203.08913},
      archivePrefix={arXiv},
      primaryClass={cs.LG},
      url={https://arxiv.org/abs/2203.08913}, 
}

@misc{rae2019compressivetransformerslongrangesequence,
      title={Compressive Transformers for Long-Range Sequence Modelling},
      author={Jack W. Rae and Anna Potapenko and Siddhant M. Jayakumar and Timothy P. Lillicrap},
      year={2019},
      eprint={1911.05507},
      archivePrefix={arXiv},
      primaryClass={cs.LG},
      url={https://arxiv.org/abs/1911.05507}, 
}

@misc{bulatov2022recurrentmemorytransformer,
      title={Recurrent Memory Transformer},
      author={Aydar Bulatov and Yuri Kuratov and Mikhail S. Burtsev},
      year={2022},
      eprint={2207.06881},
      archivePrefix={arXiv},
      primaryClass={cs.CL},
      url={https://arxiv.org/abs/2207.06881}, 
}

@misc{packer2024memgptllmsoperatingsystems,
      title={MemGPT: Towards LLMs as Operating Systems},
      author={Charles Packer and Sarah Wooders and Kevin Lin and Vivian Fang and Shishir G. Patil and Ion Stoica and Joseph E. Gonzalez},
      year={2024},
      eprint={2310.08560},
      archivePrefix={arXiv},
      primaryClass={cs.AI},
      url={https://arxiv.org/abs/2310.08560}, 
}

@misc{yen2024longcontextlanguagemodelingparallel,
      title={Long-Context Language Modeling with Parallel Context Encoding},
      author={Howard Yen and Tianyu Gao and Danqi Chen},
      year={2024},
      eprint={2402.16617},
      archivePrefix={arXiv},
      primaryClass={cs.CL},
      url={https://arxiv.org/abs/2402.16617}, 
}

@misc{jiang2023llmlingua,
  title={LLMLingua: Compressing Prompts for Accelerated Inference of Large Language Models},
  author={Huiqiang Jiang and Qianhui Wu and Chin-Yew Lin and Yuqing Yang and Lili Qiu},
  year={2023},
  eprint={2310.05736},
  archivePrefix={arXiv},
  url={https://arxiv.org/abs/2310.05736}
}

@misc{agarwal2024onpolicydistillationlanguagemodels,
      title={On-Policy Distillation of Language Models: Learning from Self-Generated Mistakes},
      author={Rishabh Agarwal and Nino Vieillard and Yongchao Zhou and Piotr Stanczyk and Sabela Ramos and Matthieu Geist and Olivier Bachem},
      year={2024},
      eprint={2306.13649},
      archivePrefix={arXiv},
      primaryClass={cs.LG},
      url={https://arxiv.org/abs/2306.13649}, 
}

@misc{feng2024alignkd,
  title={Align-KD: Distilling Cross-Modal Alignment Knowledge for Mobile Vision-Language Model},
  author={Qianhan Feng and Wenshuo Li and Tong Lin and Xinghao Chen},
  year={2024},
  eprint={2412.01282},
  archivePrefix={arXiv},
  url={https://arxiv.org/abs/2412.01282}
}

@misc{kim2024cosmos,
  title={COSMOS: Cross-Modality Self-Distillation for Vision Language Pre-training},
  author={Sanghwan Kim and Rui Xiao and Mariana-Iuliana Georgescu and Stephan Alaniz and Zeynep Akata},
  year={2024},
  eprint={2412.01814},
  archivePrefix={arXiv},
  url={https://arxiv.org/abs/2412.01814}
}

@inproceedings{naeem2024silc,
  title={SILC: Improving Vision Language Pretraining with Self-Distillation},
  author={Muhammad Ferjad Naeem and Yongqin Xian and Xiaohua Zhai and Lukas Hoyer and Luc Van Gool and Federico Tombari},
  booktitle={Computer Vision -- ECCV 2024},
  year={2024},
  doi={10.1007/978-3-031-72664-4_3},
  url={https://www.ecva.net/papers/eccv_2024/papers_ECCV/papers/03093.pdf}
}

@misc{radford2021learning,
  title={Learning Transferable Visual Models From Natural Language Supervision},
  author={Alec Radford and Jong Wook Kim and Chris Hallacy and Aditya Ramesh and Gabriel Goh and Sandhini Agarwal and Girish Sastry and Amanda Askell and Pamela Mishkin and Jack Clark and Gretchen Krueger and Ilya Sutskever},
  year={2021},
  eprint={2103.00020},
  archivePrefix={arXiv},
  url={https://arxiv.org/abs/2103.00020}
}

@misc{jia2021scalingvisualvisionlanguagerepresentation,
      title={Scaling Up Visual and Vision-Language Representation Learning With Noisy Text Supervision},
      author={Chao Jia and Yinfei Yang and Ye Xia and Yi-Ting Chen and Zarana Parekh and Hieu Pham and Quoc V. Le and Yunhsuan Sung and Zhen Li and Tom Duerig},
      year={2021},
      eprint={2102.05918},
      archivePrefix={arXiv},
      primaryClass={cs.CV},
      url={https://arxiv.org/abs/2102.05918}, 
}

@misc{alayrac2022flamingo,
  title={Flamingo: a Visual Language Model for Few-Shot Learning},
  author={Jean-Baptiste Alayrac and Jeff Donahue and Pauline Luc and Antoine Miech and Iain Barr and Yana Hasson and Karel Lenc and Arthur Mensch and Katie Millican and Malcolm Reynolds and Roman Ring and Eliza Rutherford and Serkan Cabi and Tengda Han and Zhitao Gong and Sina Samangooei and Marianne Monteiro and Jacob Menick and Sebastian Borgeaud and Andrew Brock and Aida Nematzadeh and Sahand Sharifzadeh and Mikolaj Binkowski and Ricardo Barreira and Oriol Vinyals and Andrew Zisserman and Karen Simonyan},
  year={2022},
  eprint={2204.14198},
  archivePrefix={arXiv},
  url={https://arxiv.org/abs/2204.14198}
}

@misc{li2023blip2,
  title={BLIP-2: Bootstrapping Language-Image Pre-training with Frozen Image Encoders and Large Language Models},
  author={Junnan Li and Dongxu Li and Silvio Savarese and Steven Hoi},
  year={2023},
  eprint={2301.12597},
  archivePrefix={arXiv},
  url={https://arxiv.org/abs/2301.12597}
}

@misc{kim2021ocrfree,
  title={OCR-free Document Understanding Transformer},
  author={Geewook Kim and Teakgyu Hong and Moonbin Yim and Jeongyeon Nam and Jinyoung Park and Jinyeong Yim and Wonseok Hwang and Sangdoo Yun and Dongyoon Han and Seunghyun Park},
  year={2021},
  eprint={2111.15664},
  archivePrefix={arXiv},
  url={https://arxiv.org/abs/2111.15664}
}

@misc{lee2022pix2struct,
  title={Pix2Struct: Screenshot Parsing as Pretraining for Visual Language Understanding},
  author={Kenton Lee and Mandar Joshi and Iulia Turc and Hexiang Hu and Fangyu Liu and Julian Eisenschlos and Urvashi Khandelwal and Peter Shaw and Ming-Wei Chang and Kristina Toutanova},
  year={2022},
  eprint={2210.03347},
  archivePrefix={arXiv},
  url={https://arxiv.org/abs/2210.03347}
}

@inproceedings{huang2022layoutlmv3,
  title={LayoutLMv3: Pre-training for Document AI with Unified Text and Image Masking},
  author={Huang, Yupan and Lv, Tengchao and Cui, Lei and Lu, Yutong and Wei, Furu},
  booktitle={Proceedings of the 30th ACM international conference on multimedia},
  pages={4083--4091},
  year={2022}
}

@inproceedings{li2023trocr,
  title={TrOCR: Transformer-based Optical Character Recognition with Pre-trained Models},
  author={Li, Minghao and Lv, Tengchao and Chen, Jingye and Cui, Lei and Lu, Yijuan and Florencio, Dinei and Zhang, Cha and Li, Zhoujun and Wei, Furu},
  booktitle={Proceedings of the AAAI conference on artificial intelligence},
  volume={37},
  number={11},
  pages={13094--13102},
  year={2023}
}

@misc{wei2024generalocrtheoryocr20,
      title={General OCR Theory: Towards OCR-2.0 via a Unified End-to-end Model},
      author={Haoran Wei and Chenglong Liu and Jinyue Chen and Jia Wang and Lingyu Kong and Yanming Xu and Zheng Ge and Liang Zhao and Jianjian Sun and Yuang Peng and Chunrui Han and Xiangyu Zhang},
      year={2024},
      eprint={2409.01704},
      archivePrefix={arXiv},
      primaryClass={cs.CV},
      url={https://arxiv.org/abs/2409.01704}, 
}

@article{rafailov2023direct,
  title={Direct Preference Optimization: Your Language Model is Secretly a Reward Model},
  author={Rafailov, Rafael and Sharma, Archit and Mitchell, Eric and Manning, Christopher D and Ermon, Stefano and Finn, Chelsea},
  journal={Advances in neural information processing systems},
  volume={36},
  pages={53728--53741},
  year={2023}
}

@misc{hong2024orpomonolithicpreferenceoptimization,
      title={ORPO: Monolithic Preference Optimization without Reference Model},
      author={Jiwoo Hong and Noah Lee and James Thorne},
      year={2024},
      eprint={2403.07691},
      archivePrefix={arXiv},
      primaryClass={cs.CL},
      url={https://arxiv.org/abs/2403.07691}, 
}

@article{meng2024simpo,
  title={SimPO: Simple Preference Optimization with a Reference-Free Reward},
  author={Meng, Yu and Xia, Mengzhou and Chen, Danqi},
  journal={Advances in Neural Information Processing Systems},
  volume={37},
  pages={124198--124235},
  year={2024}
}

@misc{ethayarajh2024ktomodelalignmentprospect,
      title={KTO: Model Alignment as Prospect Theoretic Optimization},
      author={Kawin Ethayarajh and Winnie Xu and Niklas Muennighoff and Dan Jurafsky and Douwe Kiela},
      year={2024},
      eprint={2402.01306},
      archivePrefix={arXiv},
      primaryClass={cs.LG},
      url={https://arxiv.org/abs/2402.01306}, 
}

@misc{wang2024mdpoconditionalpreferenceoptimization,
      title={mDPO: Conditional Preference Optimization for Multimodal Large Language Models},
      author={Fei Wang and Wenxuan Zhou and James Y. Huang and Nan Xu and Sheng Zhang and Hoifung Poon and Muhao Chen},
      year={2024},
      eprint={2406.11839},
      archivePrefix={arXiv},
      primaryClass={cs.CV},
      url={https://arxiv.org/abs/2406.11839}, 
}

@misc{jiang2025modalityfairpreferenceoptimizationtrustworthy,
      title={Modality-Fair Preference Optimization for Trustworthy MLLM Alignment},
      author={Songtao Jiang and Yan Zhang and Ruizhe Chen and Tianxiang Hu and Yeying Jin and Qinglin He and Yang Feng and Jian Wu and Zuozhu Liu},
      year={2025},
      eprint={2410.15334},
      archivePrefix={arXiv},
      primaryClass={cs.CV},
      url={https://arxiv.org/abs/2410.15334}, 
}

@misc{fu2025chipcrossmodalhierarchicaldirect,
      title={CHiP: Cross-modal Hierarchical Direct Preference Optimization},
      author={Jinlan Fu and Shenzhen Huangfu and Hao Fei and Xiaoyu Shen and Bryan Hooi and Xipeng Qiu and See-Kiong Ng},
      year={2025},
      eprint={2501.16629},
      archivePrefix={arXiv},
      primaryClass={cs.CL},
      url={https://arxiv.org/abs/2501.16629}, 
}

@misc{hsieh2024rulerwhatsrealcontext,
      title={RULER: What's the Real Context Size of Your Long-Context Language Models?},
      author={Cheng-Ping Hsieh and Simeng Sun and Samuel Kriman and Shantanu Acharya and Dima Rekesh and Fei Jia and Yang Zhang and Boris Ginsburg},
      year={2024},
      eprint={2404.06654},
      archivePrefix={arXiv},
      primaryClass={cs.CL},
      url={https://arxiv.org/abs/2404.06654}, 
}

@misc{bai2024longbenchbilingualmultitaskbenchmark,
      title={LongBench: A Bilingual, Multitask Benchmark for Long Context Understanding},
      author={Yushi Bai and Xin Lv and Jiajie Zhang and Hongchang Lyu and Jiankai Tang and Zhidian Huang and Zhengxiao Du and Xiao Liu and Aohan Zeng and Lei Hou and Yuxiao Dong and Jie Tang and Juanzi Li},
      year={2024},
      eprint={2308.14508},
      archivePrefix={arXiv},
      primaryClass={cs.CL},
      url={https://arxiv.org/abs/2308.14508}, 
}

@misc{joshi2017triviaqalargescaledistantly,
      title={TriviaQA: A Large Scale Distantly Supervised Challenge Dataset for Reading Comprehension}, 
      author={Mandar Joshi and Eunsol Choi and Daniel S. Weld and Luke Zettlemoyer},
      year={2017},
      eprint={1705.03551},
      archivePrefix={arXiv},
      primaryClass={cs.CL},
      url={https://arxiv.org/abs/1705.03551}, 
}

@inproceedings{ho-etal-2020-constructing,
    title = "Constructing A Multi-hop {QA} Dataset for Comprehensive Evaluation of Reasoning Steps",
    author = "Ho, Xanh  and
      Duong Nguyen, Anh-Khoa  and
      Sugawara, Saku  and
      Aizawa, Akiko",
    editor = "Scott, Donia  and
      Bel, Nuria  and
      Zong, Chengqing",
    booktitle = "Proceedings of the 28th International Conference on Computational Linguistics",
    month = dec,
    year = "2020",
    address = "Barcelona, Spain (Online)",
    publisher = "International Committee on Computational Linguistics",
    url = "https://aclanthology.org/2020.coling-main.580/",
    doi = "10.18653/v1/2020.coling-main.580",
    pages = "6609--6625"
}

@misc{cobbe2021trainingverifierssolvemath,
      title={Training Verifiers to Solve Math Word Problems}, 
      author={Karl Cobbe and Vineet Kosaraju and Mohammad Bavarian and Mark Chen and Heewoo Jun and Lukasz Kaiser and Matthias Plappert and Jerry Tworek and Jacob Hilton and Reiichiro Nakano and Christopher Hesse and John Schulman},
      year={2021},
      eprint={2110.14168},
      archivePrefix={arXiv},
      primaryClass={cs.LG},
      url={https://arxiv.org/abs/2110.14168}, 
}

\appendix

\end{document}